\documentclass[conference]{IEEEtran}
\IEEEoverridecommandlockouts
\usepackage{cite}
\usepackage{amsmath,amssymb,amsfonts}
\usepackage{algorithm}
\usepackage{algpseudocode}  
\usepackage{graphicx}
\usepackage{textcomp}
\usepackage[table,xcdraw]{xcolor}
\usepackage{seqsplit}
\usepackage{setspace}
\usepackage{changes}
\usepackage{hyperref}
\def\BibTeX{{\rm B\kern-.05em{\sc i\kern-.025em b}\kern-.08em
    T\kern-.1667em\lower.7ex\hbox{E}\kern-.125emX}}

\begin{document}

\title{ExScene: Free-View 3D Scene Reconstruction with Gaussian Splatting from a Single Image
\thanks{\dag: Equal Contribution. 

\ddag: CorrespondingAuthor: Fangxin Wang.}
}
\DeclareRobustCommand*{\IEEEauthorrefmark}[1]{%
    \raisebox{0pt}[0pt][0pt]{\textsuperscript{\footnotesize\ensuremath{#1}}}}

	\author{
    
		\IEEEauthorblockN{
			Tianyi Gong\IEEEauthorrefmark{1,2\dag}, 
			Boyan Li\IEEEauthorrefmark{1,2\dag},
			Yifei Zhong\IEEEauthorrefmark{1,2},
			Fangxin Wang\IEEEauthorrefmark{2,1\ddag}}
		\IEEEauthorblockA{
        \IEEEauthorrefmark{1}Shenzhen Future Network of Intelligence Institute
        \\
        \IEEEauthorrefmark{2}School of Science and Engineering, The Chinese University of Hongkong, Shenzhen
			\\
            		Email: \{tianyigong1, 
                        boyanli, 
                        yifeizhong\}@link.cuhk.edu.cn,
                        wangfangxin@cuhk.edu.cn
		}
	}

\maketitle
\begin{figure}
    \centering
    \includegraphics[width=1\linewidth]{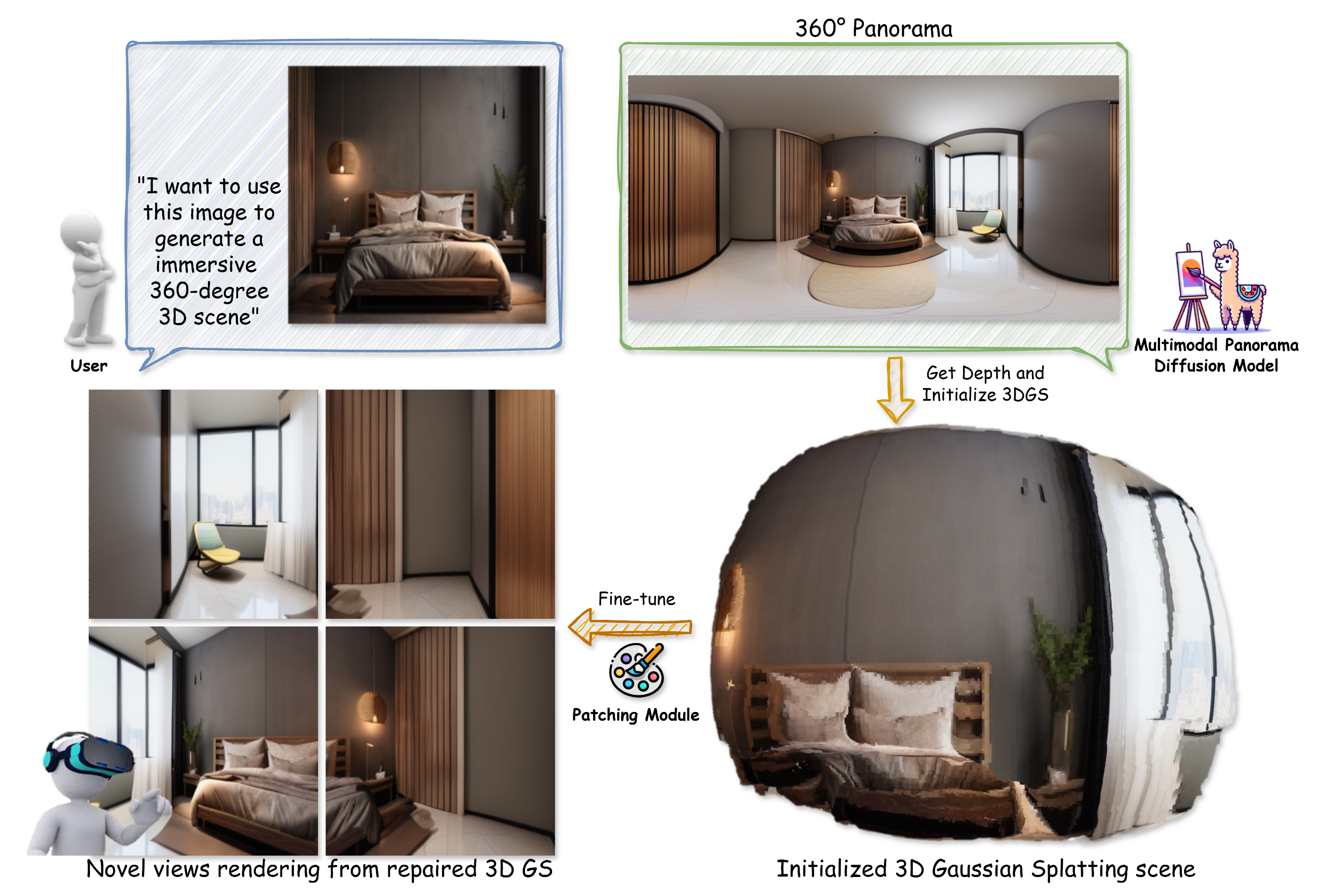}
    \caption{\textbf{ExScene} is a novel reconstruction method that extends and reconstructs any given single narrow-view image into an immersive 360-degree 3D scene based on 3D Gaussian Splatting (3DGS).}
    \label{fig:intro}
\end{figure}

\begin{abstract}
The increasing demand for augmented and virtual reality applications has highlighted the importance of crafting immersive 3D scenes from a simple single-view image. However, due to the partial priors provided by single-view input, existing methods are often limited to reconstruct low-consistency 3D scenes with narrow fields of view from single-view input. These limitations make them less capable of generalizing to reconstruct immersive scenes. To address this problem, we propose ExScene, a two-stage pipeline to reconstruct an immersive 3D scene from any given single-view image. ExScene designs a novel multimodal diffusion model to generate a high-fidelity and globally consistent panoramic image. We then develop a panoramic depth estimation approach to calculate geometric information from panorama, and we combine geometric information with high-fidelity panoramic image to train an initial 3D Gaussian Splatting (3DGS) model. Following this, we introduce a GS refinement technique with 2D stable video diffusion priors. We add camera trajectory consistency and color-geometric priors into the denoising process of diffusion to improve color and spatial consistency across image sequences. These refined sequences are then used to fine-tune the initial 3DGS model, leading to better reconstruction quality. Experimental results demonstrate that our ExScene achieves consistent and immersive scene reconstruction using only single-view input, significantly surpassing state-of-the-art baselines.
\end{abstract}


\section{Introduction}

\label{sec:intro}
Single-view 3D reconstruction technology is crucial for computer vision, robotics, and augmented reality/virtual reality (AR/VR) \cite{wang2024diffusion}. However, challenges such as limited scene priors, poor scene consistency in reconstruction, and artifacts and hollow regions significantly limit their applicability in real-world scenarios. Although neural radiance fields (NeRF) \cite{mildenhall2021nerf} and 3D Gaussian splatting (3DGS) \cite{kerbl20233d} have advanced novel view synthesis and 3D scene building but rely on numerous sequential images taken in controlled trajectories, which is an impractical requirement for real-world applications. While recent breakthroughs in diffusion models \cite{ho2020denoising} \cite{sauer2025adversarial} and autoregressive models \cite{li2024autoregressive} have enabled remarkable progress in single-view reconstruction, most existing methods \cite{liu2024novel} \cite{wang2024vistadream} \cite{yu2024viewcrafter} are primarily restricted to front-facing scenarios and are limited to enabling small-angle rotations or minor viewpoint extrapolations. They fail to generate immersive 360-degree reconstructions due to errors in iterative processes, introducing geometric distortions, inconsistent colors, and artifacts. In short, generating immersive, high-fidelity 3D reconstructions from a single view remains a significant unsolved challenge. 

To address these challenges, we introduce ExScene. As illustrated in Fig.~\ref{fig:intro}, our method innovatively fuses multimodal panoramic image generation technique and the 3DGS scene refinement approach to achieve high-fidelity color representation and geometric consistency in 3D reconstructions. First, we design a multimodal diffusion model with panoramic priors to address semantic inconsistencies in panoramic image generation. This special model fuses text and image features into the denoising process to generate semantically consistent high-fidelity panoramas. Subsequently, we propose an improved panoramic depth estimation method to overcome projection distortions in exiting panoramic depth estimation methods. Finally, we used the precise geometric information and high-fidelity panoramic image to train an initial 3DGS model. In the second stage, we introduce a special Gaussian patching module to address viewpoint occlusion artifacts in initial 3D model. Our patching module is driven by a particular stable video diffusion (SVD) model, which fuses multi-view dimension-aware capabilities and color-geometry priors. This ensures 3D-view consistency across the generated 2D image sequence, improving the 3DGS model's quality.

Our experiments on diverse single-view datasets show that ExScene surpasses state-of-the-art methods qualitatively and quantitatively. Ablation studies confirm the effectiveness of our multimodal panoramic image generation module and Gaussian patching module in generating high-fidelity panoramic images and reconstructing high-quality and consistent 3D scenes. The contributions of this paper are summarized as follows:

\begin{itemize}
    \item We design ExScene. To our best knowledge, we are the first to design a framework for extending and reconstructing a single-view image to a free-view 3D scene.

    \item We propose a multimodal panoramic diffusion model that effectively addresses the issues of geometric and semantic inconsistencies in panoramic image generation encountered by existing methods, improving the quality of panoramic images generated from single-view input.

    \item To improve reconstruction quality, we introduce a novel SVD model with multiple priors to optimize the video sequences rendered from the initial model. The optimized images are then used to fine-tune the 3D Gaussians, resulting in high-quality and consistent 3D scenes.
\end{itemize}

\section{Related Work}
\label{sec:Related work}
\subsection{Panoramic image generation}
Panoramic images provide more 3D scene details than traditional 2D images. MVDiffusion\cite{Tang2023mvdiffusion}
fine-tunes a text-to-image diffusion model to create high-resolution panoramas from eight input perspectives. Meanwhile, artifacts often appear in the "sky" and "floor" regions. Diffusion360 \cite{feng2023diffusion360} is capable of generating high-resolution panoramic images using a circular blending strategy to maintain consistency but it also introduces additional unwanted image details.

\subsection{Single-view reconstruction}
Single-view reconstruction mainly focuses on generating a 3D scene from a single image to facilitate the rendering of novel views \cite{feng2023diffusion360}.
Several methods aim to convert single-view images into 3D objects through end-to-end training, producing various 3D representations, such as multiview-consistent images \cite{li2024era3d}, meshes\cite{liu2024one}, neural fields \cite{song2020denoising}, and 3D Gaussian Splatting\cite{gao2024cat3d} \cite{shriram2024realmdreamer}. On the one hand RealmDreamer \cite{shriram2024realmdreamer} leverages pre-trained 2D inpainting to generate 3D scenes, but relies on text prompts. On the other hand, Cat3D\cite{gao2024cat3d} utilize a MD diffusion model to create novel views. But these novel views struggle with 3D consistency.

\begin{figure*}
    \centering
    \includegraphics[width=1\linewidth]{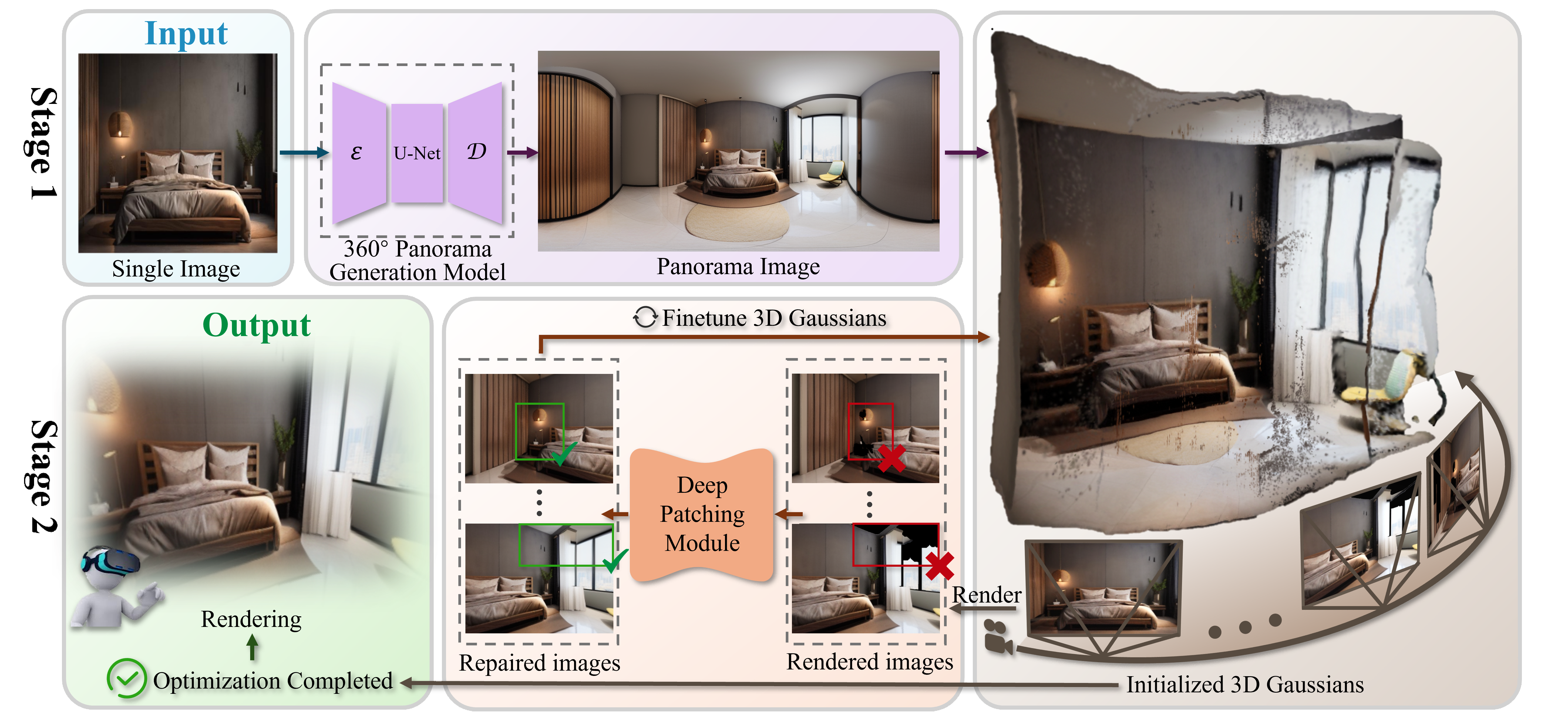}
    \caption{\textbf{Overview of ExScene.} \textit{Stage 1: Generate initialized Gaussian scenes.} Given a single-view image as input, ExScene introduces a 2D multimodal diffusion model with panoramic priors to generate a high-quality and consistent panoramic image (Section~\ref{subsec:A}). Subsequently, ExScene designs a depth estimation module with distortion-elimination regularization to calculate geometric parameters of the generated panorama (Section~\ref{subsec:B}). The depth map is then combined with the panoramic image to train an initial 3DGS (Section~\ref{subsec:C}). \textit{Stage 2: Fine-tune initialized 3D Gaussians.} To address the missing regions and geometric distortions present in the initial 3DGS, we design an innovative Patching Module that consists of an SVD model fusing camera trajectory consistency and image color-geometric priors. Firstly, ExScene simulates virtual camera trajectories over the initial 3DGS to obtain rendered image sequences. These sequences are then refined using the patching module, resulting in high-fidelity and consistent repaired views. Finally, we use these repaired sequences to fine-tune initial 3DGS, improving the reconstruction quality (Section~\ref{subsec:D}).}
    \label{fig:overview}
\end{figure*}

\section{Method}
\label{sec:Method}
In this section, we provide a detailed overview of our proposed ExScene. As illustrated in Fig.\ref{fig:overview}, our approach consists of two key stages: \textit{Generate initialized Gaussian scenes} and \textit{Fine-tune initialized 3D Gaussians.}
\subsection{Single-View Image to Panoramas}
\label{subsec:A}
To generate a panoramic image from a single input image can effectively provide an overview of the entire scene, essential for creating consistent 360-degree immersive 3D scenes.

\textit{Immersive Panoramic Image Generation.} We design a novel latent diffusion model with multi-feature priors to generate a panoramic image with color and semantic consistency from a single-view image input. This model uses a Variational Autoencoder (VAE) to convert Gaussian noise to latent features. We then propose multimodal features $F_{multi}$ as conditional inputs for U-Net to perform multiple denoising iterations, aiming to achieve the desired information distribution. Finally, we used the VAE to decode the latent space and generate the final image. This multi-step denoising process in the latent space simplifies the training procedure, while the incorporated multimodal information effectively addresses the semantic and geometric inconsistencies present in current panorama generation methods, enhancing image quality.

For the multimodal feature $F_{multi}$, we use a pre-trained LLaVA \cite{liu2024visual} model to generate semantic descriptions of the image. Then, we design a special multimodal encoder to extract semantic and visual features from the text description and input image. Finally, we combine the textual feature $F_{text}$ and visual feature $F_{vision}$ to create the modality-fused conditional input $F_{multi}$, defined as:
 \begin{equation}F_{\text{multi}} = F_{\text{text}} \oplus F_{\text{vision}}\end{equation}

\subsection{Smooth-Guided Panoramic Depth Initialization}
\label{subsec:B}
\textit{Panoramic Image Depth Estimation.} We introduce a novel panoramic depth estimation approach with distortion-elimination regularization to address projection distortion in calculating geometric information for panoramic images. Inspired by 360MonoDepth \cite{DBLP:journals/corr/abs-2111-15669}, for a panorama image \(I_{rgb}\), we first project the image onto \(K\) planes, obtaining a set of images \(\{\overline{I}^k_{rgb}\}^K_{k=1}\) . We set \(K=20\), as our experiments indicate that an icosahedron provides a uniform coverage of the spherical surface. After obtaining the projected images, we apply monocular depth estimation to generate the depth maps \(\{\overline{I}^K_d\}\) for each projected image. Since these maps are independently predicted, they have different scales and offset.To ensure depth consistency in the panoramic depth map, we optimize an affine transformation field across multiple viewpoints. For each depth \(\{\overline{I}^k_d=D^k(\theta, \phi)\}\), we create an affine transformation field that includes spatial scale \(s(\theta, \phi)\) and offset \(o(\theta, \phi)\). Let \((\theta, \phi)\) be the coordinates of each pixel, and the affine operation for each pixel can be represented as:
\begin{equation}
\tilde{D}^k_{theta, \phi}=s(\theta, \phi)\cdot D^k(\theta, \phi)+o(\theta, \phi)
\end{equation}
To align all depth maps of the planes, we optimize the affine transformation field to minimize the energy function, which ensures that the depth maps in overlapping regions of all planes are consistent. The energy function is described as:
\begin{equation}
   E=\sum_{(a,b)\leq K}\sum_{(\theta, \phi)\in \Omega(a,b)} \tilde{D}_1*(\theta, \phi)-\tilde{D}_b(\theta, \phi)+E_{reg} 
\end{equation}
where $E_{\text{reg}}$ is the regularization energy function, which will be introduced in the following paragraph.
\begin{spacing}{1.0}
\textit{Distortion-Elimination Regularization.} We observe that perspective distortion can easily occur during the affine transformation process in multiple views. To address this, we apply a series of regularization processes to ensure the continuity of optimization. Firstly, the continuity term encourages smooth changes in scale and offset between adjacent grid points, preventing unnatural jumps or abrupt fluctuations in depth map during the adjustment process. Secondly, the scale regularization term constrains the scale values close to $1$. Lastly, The regularization term limits the absolute values of affine scale and offset, suppressing excessive fluctuations, ensuring stability in the optimization process, and preventing distortion effects caused by unreasonable parameter adjustments. The regularization energy function is defined as follows: 
\end{spacing}
{\small
\[
  E_{reg}=\sum_{a,b\leq K} \sum_{(\theta,\phi)\in \Omega}(s_a(\theta, \phi)-s_b(\theta, \phi))^2+(o_a(\theta, \phi)-o_b(\theta, \phi))^2
\]
\begin{equation}
    +\sum_{\theta, \phi \in \Omega}(s(\theta, \phi)-1)^2 + \sum_{(\theta, \phi)\in \Omega}(|s(\theta, \phi)|+|o(\theta, \phi)|)
\end{equation}
}
where \(s(\theta, \phi)\) and \(o(\theta, \phi)\) is the scale factor 
and the offset factor at pixel position \(\theta, \phi\). \(\Omega\) is the region of a plane, \(\Omega\) is the overlapping region of plane \(a\) and plane \(b\).

By combining these three optimization objectives, we effectively ensure the accuracy and stability of the affine alignment of the depth map, allowing the transformation process to effectively preserve the smoothness and consistency of the depth information, avoiding distortion caused by over-adjustment.

\subsection{Multi-Aware Regularized Gaussian Reconstruction}
\label{subsec:C}
\textit{Initial Gaussian Reconstruction.} In this step, we reconstruct an initial 3D GS model to provide a global structural prior for the scene. The initial 3D model is crucial for the final reconstruction quality, as it provides spatial and structural information to mitigate spatial distortion and geometric deformation. However, occlusion biases inherent to panoramic perspectives and inaccuracies in panoramic depth estimation result in holes and artifacts in the reconstructed 3DGS. To address this issue, we optimize the Gaussian iteration process by combining the original photometric loss with our proposed geometry-aware loss, enhancing the optimization and minimizing the impact of depth inaccuracies and view occlusion.

\textit{Photo-metric Loss.} Aligned with original 3DGS, we initially compute the photo-metric loss between the input RGB images and Gaussian-rendered images. The photo-metric loss function combines $L_{1}$ with an SSIM term $L_{SSIM}$.
\begin{equation}
    L_{pho}=(1-\lambda_{pho} )L_1+\lambda L_{SSIM}
\end{equation}
where $\lambda_{pho}$ represents a hyperparameter, and $L_{pho}$ denotes the photo-metric loss.

\textit{Geometry-Aware Regulization.} Although depth provides distance information within the scene, abrupt changes in depth gradients can lead to geometric distortions in the reconstructed model. To address this issue, we propose a joint regularization method that combines surface normal priors and depth gradients to constrain the training process. Specifically, we align local depth gradients with surface normals through vector multiplication, enforcing their consistency at each surface point. Through this method, we ensure geometric realism in the reconstructed scene, effectively avoiding artifacts from distortions and unnatural transitions. The regularization formulation is expressed as follows:
\begin{equation}L_{geo}=\sum_{i,j}||(\nabla D_{i,j}) \cdot \hat{N}_{i,j}||\end{equation}
where $\nabla D_{i,j}$ present depth gradient, and $\hat{N}_{i,j}$ present normal vector in the location.

\textit{Multi-Loss Optimization.} Finally, during Gaussian reconstruction process, we design a multi-loss optimization function to guide and constrain the Gaussian iteration. Alongside photometric loss, the introduced normal-depth gradient loss effectively resolves geometric distortions and artifacts.
\begin{equation}L_{gaussian}=L_{pho}+\lambda_{geo}L_{geo}\end{equation}
where $L_{gaussian}$ is the loss function used in our Gaussian reconstruction stage. The weights $\lambda_{geo}$ serve as hyperparameters to adjust the influences of geometry-aware regulization.

\subsection{Fine-Tuning Initial Gaussian with SVD}
\label{subsec:D}
In the previous stage, we reconstructed an initial 3DGS model using depth estimation and a novel loss function. However, due to the viewpoint limitations of single images and depth estimation errors, this 3DGS model exhibited many artifacts and blurry impressions. In this stage, we use a novel iterative optimization method to enhance scene details. We planned a virtual viewpoint trajectory to maximize exposure of artifacts and missing areas. Subsequently, we used a patching module that incorporates Stable Video Diffusion (SVD) priors and color-geometric priors to repair Gaussian rendering image sequences from virtual viewpoints, transforming incomplete images into clear, realistic representations. After enhancement, these refined image sequences serve as supplemental guidance for fine-tuning the initial Gaussian model, which aids in effectively optimizing the Gaussian kernels in combining with previous normal regularization and geometric regularization. This method reduces holes, distortions, and blurry artifacts in the Gaussian model, enhancing overall quality.

\textit{Multidimension-Constrained SVD.} To ensure that the repaired 2D video frames maintain high-quality and consistent outputs along a specific camera path, we propose a special trajectory-constrained video diffusion method that leverages viewpoint trajectory priors to enhance the global consistency in video generation. Specifically, we adopt the Diffusion Transformers (DiT) network \cite{peebles2023scalable}, based on ViT (Vision Transformer) \cite{dosovitskiy2020image}, as the backbone of our diffusion model and we train it on a novel view synthesis dataset \cite{liu20243dgs}, where the learned priors effectively guide the 2D video diffusion process. We render a video sequence $V_{\text{render}} = \{I_{1}^{\text{render}}, \dots, I_{L}^{\text{render}}\}$ along a predefined trajectory on the 3D GS model, serving as the spatial representation input. To address distortions in long video outputs often caused by SVD models, we perform virtual viewpoint video rendering with an extremely low frame rate. This video sequence is processed by a VAE model to extract latent features, which are then tokenized into token sequences $T = \{T_{1}, \dots, I_{L}\}$ for the DiT network. In contrast to traditional SVD models, which extract single-image features as conditional inputs, we input $V_{\text{render}}$ into a CLIP encoder \cite{radford2021learning} to compute implicit encoding vectors $\mathbf{c}_{render}$. These vectors served as conditional inputs for the DiT model, ensuring 3D consistency during the diffusion process. We introduce an additional multi-head cross-attention layer in the DiT blocks to enhance the interaction between conditional inputs and latent features. 
To fuse color information and spatial priors from latent features, 
we input the single-view input image into a CLIP image encoder to obtain the encoding vector $\mathbf{c}_{input}$. The vector modulates the features in the DiT network via cross-attention, ensuring consistent color styles in the repaired frames. Our video diffusion model predicts noise at each time step. The training objective is defined as follows: 
\begin{equation}\min _{\theta}=\mathbb{E}_{t \sim \mathcal{U}(0,1), \epsilon \sim \mathcal{N}(\mathbf{0}, \boldsymbol{I})}\left[\left\|\epsilon_{\theta}\left(\boldsymbol{x}_{t}, t, \mathbf{c}\right)-\epsilon\right\|_{2}^{2}\right]\end{equation}
where $\boldsymbol{x}_{t}=\alpha_{t} \boldsymbol{x}_{0}+\sigma_{t} \epsilon$,  hyperparameters $\alpha_{t}$ and $\sigma_{t}$ characterize the noise at time step $t$, and $\mathbf{c}$ represents the summarize of $\mathbf{c}_{render}$ and $\mathbf{c}_{input}$.

With the assistance of the trained DiT network, the noisy latent representations are iteratively denoised into clean latent representations $I_{repair} = \{I_{1}^{\text{repair}}, \dots, I_{L}^{\text{repair}}\}$, which are then decoded into high-quality novel views using a VAE decoder.
\begin{figure*}
    \centering
    \includegraphics[width=0.985\linewidth]{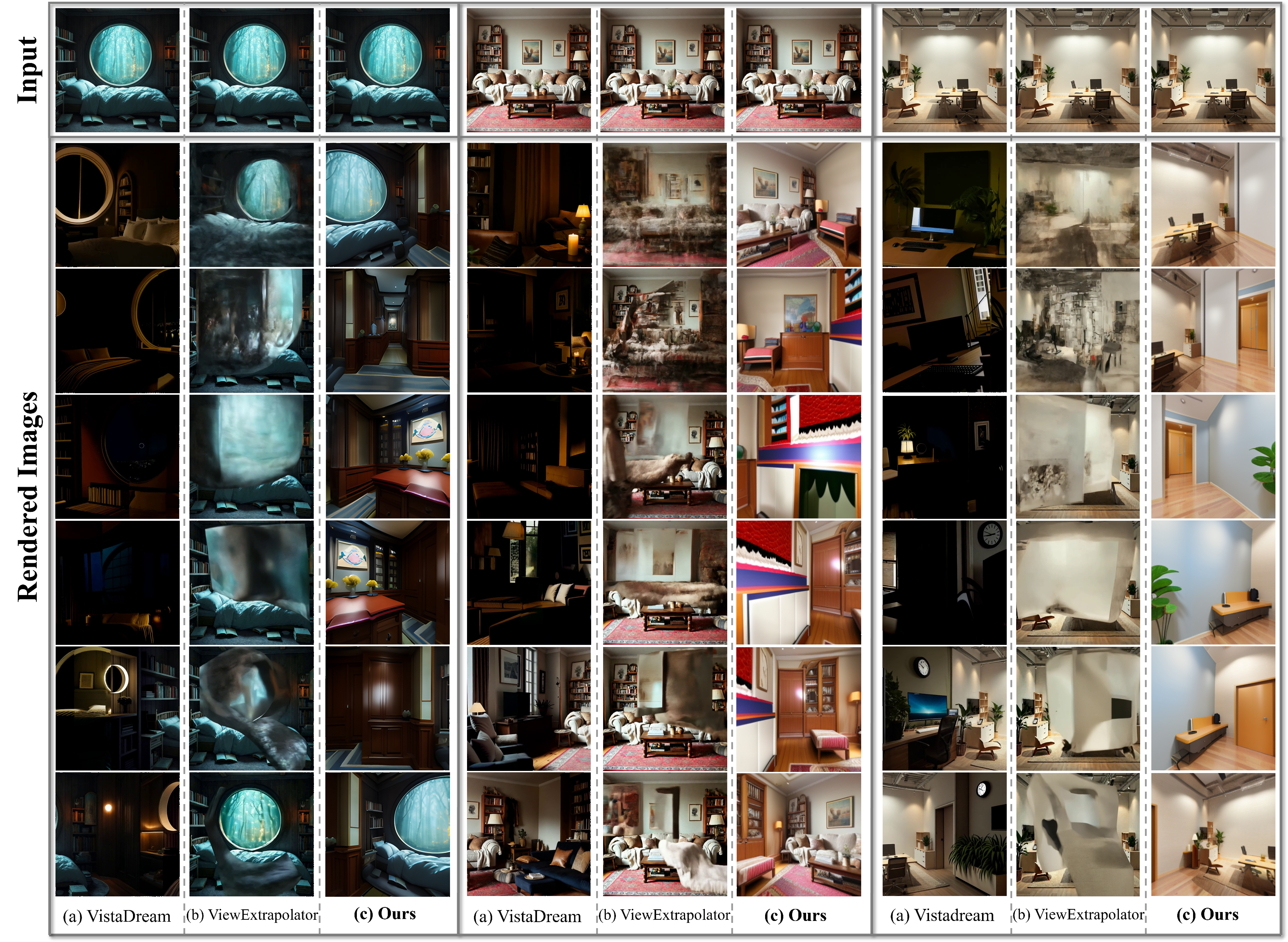}
    \caption{\textbf{Qualitative comparison of our method with ViewExtrapolator \cite{liu2024novel} and VistaDream \cite{wang2024vistadream}.} Given a single-view input image, (a) Vistadream first outpaints the image and reconstructs a Gaussian scaffold, then uses a consistent diffusion model to fine-tune the scaffold. (b) ViewExtrapolator employs SVD model generation priors to achieve realistic new view extrapolation. (c) ExScene (Ours) design a two-stage framework for single-view reconstruction.}
    \label{fig:result}
\end{figure*}

\begin{table*}[htbp]
    \centering
    \caption{Quantitative evaluations for rendered images from the reconstructed scenes.}
    \vspace{-2.5mm}
    \begin{tabular}{cccccccccc}
\noalign{\hrule height 1.25pt}  
                 Method & Q-Align $\uparrow$ & Noisy-free $\uparrow$ & Edge $\uparrow$ & Structure $\uparrow$ & Detail $\uparrow$ & Quality $\uparrow$ & Light $\uparrow$ & Color $\uparrow$\\ \hline
                ViewExtraPolator & 3.049 & 0.471 & 0.437 & 0.384 & 0.714 & 0.244 & 0.537 & 0.588\\
                Vistadream       & 3.251 & 0.638 & 0.593 & 0.628 & 0.725 & 0.400 & 0.418 & 0.810\\
\rowcolor{gray!15} w/o Patching Module  & 3.317 & 0.683 & 0.673 & 0.552 & 0.730 & 0.450 & 0.517 & 0.670\\
\rowcolor{gray!30} \textbf{Ours} & \textbf{3.412} & \textbf{0.790} & \textbf{0.689} & \textbf{0.684} & \textbf{0.736} & \textbf{0.642} & \textbf{0.654} & \textbf{0.824}\\
\noalign{\hrule height 1.25pt}  
    \end{tabular}
    \label{tab:result}
\end{table*}

\section{Experiments}
\label{sec:Experiments}
\subsection{Experimental protocol}
\textbf{\textit{Datasets.}}
Currently, there is no publicly available unified image dataset for single-view reconstruction of immersive 3D scenes. Therefore, we used 62 single-view images from baseline methods \cite{wang2024vistadream}, as well as generated using Stable Diffusion V2.1 \cite{rombach2022high}, as test images for ExScene and other methods, covering various indoor and outdoor, real and simulated scenes.

\textbf{\textit{Implementation.}}
We conducted experiments with 4 Nvidia 3090 GPUs. For a single-view input, we used LLaVA \cite{liu2024visual} to generate an image description. The image and description are incorporated into the denoising process of the diffusion model to generate panoramic images. We trained SVD model on a novel view synthesis dataset \cite{liu20243dgs} to learn trajectory priors.

\textbf{\textit{Baselines.}}
We compare our method against 2 SOTA baseline approaches:  ViewExtrapolator \cite {liu2024novel}  and  VistaDream \cite{wang2024vistadream}. ViewExtrapolator utilizes a redesigned SVD denoising process and generative priors for image sequence artifact repair.
VistaDream uses a diffusion model to refine a 3D scaffold and improve view consistency through multi-view sampling.


\textbf{\textit{Evaluation Metrics.}}
Following previous work \cite{wang2024vistadream} \cite{zhou2025dreamscene360}, we use no-reference image quality metrics. QAlign \cite{wu2023q} assesses image quality using a model fine-tuned on image quality datasets. For novel view synthesis, we employ GPT-4o model to evaluate multi-view images in seven aspects: noise level, edge smoothness, structural integrity, image detail, overall quality, light consistency, and color consistency. 

\subsection{Comparisons with baselines}
\textit{Main Results.} In Fig.~\ref{fig:result}, we present the qualitative results of our method compared to baseline methods evaluated on our single-view image dataset. Due to error accumulation, VistaDream exhibits color and style incosisitencies as illustrated in column (a) of Fig.~\ref{fig:result}. In column (b) of Fig.~\ref{fig:result}, the absence of camera trajectory priors in ViewExtrapolator's SVD module leads to a loss of 3D consistency in the synthesized novel views, causing distortions and artifacts in the rendered images.

The quantitative results in TABLE~\ref{tab:result} and qualitative examples in Fig.~\ref{fig:result} both demonstrate that our proposed ExScene, using a multimodal panoramic diffusion model, significantly outperforms existing methods in expanding scene representation. This enhancement provides panoramic geometric and color priors for initializing and fine-tuning the subsequent 3D Gaussian model, ensuring high-quality 3D scene reconstruction. Furthermore, the patching model ensures multi-view consistency during the scene optimization while simultaneously improving overall reconstruction quality. This ultimately results in accurate and photorealistic scene reconstruction.

\begin{table}[htbp]
    \small
    \centering
    \caption{Ablation Study Quantitative Result of Patching Module.}
    \vspace{-2.5mm}
    \begin{tabular}{ccccc}
\noalign{\hrule height 1.25pt}  
           Method & Edge $\uparrow$ & Structure $\uparrow$ & Light $\uparrow$ & Color $\uparrow$\\ \hline
           w/o Text Feature  & 0.393 & 0.405 & 0.413 & 0.781\\
\rowcolor{gray!30} \textbf{Ours} & \textbf{0.467} & \textbf{0.445} & \textbf{0.467} & \textbf{0.820}\\
\noalign{\hrule height 1.25pt}  
    \end{tabular}
    \label{tab:ablation}
\end{table}

\subsection{Ablation Study}
\textit{The Importance of Text Feature in Panorama Generation.} We removed the LLaVA-based description generation module, as shown in TABLE~\ref{tab:ablation}, we evaluate the generated images using consistency metrics. Images generated without semantic guidance show jagged edges and inconsistent lighting and color, highlighting the effectiveness of our semantic encoding.

\textit{The Importance of the Gaussian Patching Module.} The pathcing module is essential for repairing holes and eliminating artifacts, significantly impacting the final quality.  When the module was removed, as shown in TABLE~\ref{tab:result} w/o Patching Module, the quality of the Gaussian model significantly degraded after losing the second-stage fine-tuning process. In contrast, our patching module effectively eliminates distortions, enhancing the realism and quality of the scenes.


\section{Conclusion}
\label{sec:Conclusion}
In this paper, we introduce ExScene, a two-stage framework for generating immersive 360-degree 3D scenes from a single-view image. In the first stage, we employ a multimodal diffusion model with panoramic priors to generate high-quality panoramic images, along with a panoramic depth estimation module to predict depth information. These are then combined to train an initial 3D Gaussians model. In the second stage, we propose a dimension-aware stable video diffusion method to repair video rendered from the initial 3D scene, producing high-quality and consistent multi-view images. These images are then used to refine the initial model to achieve realistic visual effects. Extensive experimental results demonstrate that ExScene surpasses state-of-the-art performance in single-view generation of immersive 360-degree scenes.


\section*{Acknowledgments}
\label{sec:Acknowledgments}
{\setlength{\emergencystretch}{3em}
\tolerance=2000       
\sloppy               
This work was supported in part by NSFC with Grant No. 62293482, the Basic Research Project No. \seqsplit{HZQBKCZYZ2021067} of Hetao Shenzhen-HK S\&T Cooperation Zone, NSFC with Grant No. 62471423, the Shenzhen Science and Technology Program with Grant No.  \seqsplit{JCYJ20241202124021028} and Grant No.  \seqsplit{JCYJ20230807114204010}, the Shenzhen Outstanding Talents Training Fund 202002, the Guangdong Research Projects No.  \seqsplit{20190CX01X104}, the Young Elite Scientists Sponsorship Program of CAST (Grant No.  \seqsplit{2022QNRC001}), the Guangdong Provincial Key Laboratory of Future Networks of Intelligence (Grant No.  \seqsplit{2022B1212010001}) and the Shenzhen Key Laboratory of Big Data and Artificial Intelligence (Grant No.  \seqsplit{ZDSYS201707251409055}).
}

\bibliographystyle{IEEEbib}
\bibliography{ref}
\end{document}